\documentclass[10pt,english]{article}
\usepackage{subfigure}
\usepackage{graphicx}
\usepackage{float}
\usepackage[T1]{fontenc}
\usepackage[latin9]{inputenc}
\usepackage[margin=1.25in]{geometry}
\usepackage{float}
\usepackage{amsmath}
\usepackage{amsbsy}
\usepackage{setspace}
\usepackage{amssymb}
\usepackage{esint}
\usepackage{cite}
\newfloat{algorithm}{tbp}{loa}
\floatname{algorithm}{Algorithm}
\usepackage{algorithmic}
\usepackage{hyperref}
\usepackage{listings}

\newlength\myindent
\makeatletter

\floatstyle{ruled}
\newfloat{algorithm}{tbp}{loa}
\floatname{algorithm}{Algorithm}

\usepackage{babel}

\begin{document}

\title{Diffusion Self-Organizing Map on the Hypersphere}

\author{M. Andrecut}


\maketitle
{

\centering Calgary, Alberta, T3G 5Y8, Canada

\centering mircea.andrecut@gmail.com

} 
\bigskip 
\begin{abstract}

We discuss a diffusion based implementation of the self-organizing map on the unit hypersphere. We show that this approach can be efficiently 
implemented using just linear algebra methods, we give a python numpy implementation, and we illustrate the approach using the 
well known MNIST dataset. 

\smallskip 


\end{abstract}

\section{Introduction}

The Self-Organizing Map (SOM) is an Artificial Neural Network (ANN) that can be used to learn a map that projects a high-dimensional input space 
into a low-dimensional space, by preserving the topological structure of the input space, such as the local neighborhood \cite{key-1}, \cite{key-2}. 
After the convergence of the SOM learning process, the nodes (neurons) of the network are self-organized such that input data samples associated with two nearby nodes are similar. 
The learning in SOM is based on biologically inspired competitive learning  \cite{key-3}, contrary to the more traditional types of ANNs, like the feed-forward networks, 
where learning is based on some variant of the backpropagation algorithm. Moreover, this type of competitive learning is also unsupervised, that is SOM learns from unlabeled data, like 
in the more traditional clustering methods. Due to their combined clustering and visualization properties SOMs have been successfully applied in almost every 
scientific and engineering domain \cite{key-4}. 

Today there exist several variations of the original SOM algorithm, however most of these variants are based 
on the Euclidean distance as a similarity measure \cite{key-5}. Here we discuss a version of SOM that is implemented on the unit hypersphere, and uses the dot product as a similarity measure. The nodes of the network also use a diffusion 
mechanism to communicate among them. We show that this approach can be efficiently implemented using just linear algebra methods. A python numpy implementation is also 
provided, and the method is illustrated using the MNIST dataset, which is a well known benchmark frequently used in machine learning.

\section{Self-Organizing Map}

We assume that $X=\{x^{(n)} \in \mathbb{R}^d \mid  n=0,1,...,N-1 \}$ is the input set of $N$ $d$-dimensional data points. 
A SOM consists of a number of $K$ neurons $U=\{ u^{(k)} \in \mathbb{R}^d \mid  k=0,1,...,K-1 \}$, associated with $K$ points distributed in a 2-dimensional regular lattice (rectangular or hexagonal). 

The original online SOM is a recurrent algorithm where at each time step one input data sample is presented to the network, and a competitive method is used 
to select the best matching neuron (the closest one). The neurons in the neighborhood of the "winning neuron" are updated such that they are getting closer to the input sample \cite{key-5}. 
We should note that here we have two distances, the first one is defined in the high-dimensional input space and it used to compute the "winning neuron", while the second distance is used in the 2-dimensional space in order to   
define the local neighborhood, and the "interaction" between the neurons. Using the Euclidean distance in both spaces, the online SOM algorithm can be formulated as following:
\begin{enumerate}
\item A randomly selected input $x^{(n)}(t) \in X$ is presented to the network at the time step $t$: $n=\text{rand}(0,N)$ (here, $\text{rand(0,N)}$ returns a random integer $0\leq n <N $).
\item A "winning node" $u^{(k^*)}(t) \in U$ that closely matches $x^{(n)}(t)$ is selected:
\begin{equation}
k^* = \text{arg} \min_k \Vert x^{(n)}(t) - u^{(k)}(t) \Vert, 
\end{equation}
where $\Vert \cdot \Vert$ is the Euclidean norm.
\item The weights of all nodes are then updated using the following equation:
\begin{equation}
u^{(k)}(t+1) = u^{(k)}(t) + \alpha(t) h_{k^*,k}(t)[x^{(n)}(t)-u^{(k)}(t)].
\end{equation}
\end{enumerate}
Here, $\alpha(t)>0$ is the learning rate, and $h_{k^*,k}(t)$ is the neighborhood function centered on the "winning node", at the time step $t$. 
The standard choice for $h$ is the Gaussian function:
\begin{equation}
h_{k^*,k}(t) = \exp \left( -\frac{\Vert u^{(k^*)}(t) - u^{(k)}(t) \Vert^2}{2\sigma^2(t)} \right), 
\end{equation}
where $\sigma(t)$ defines the neighborhood size at time step $t$. One can also use other neighborhood functions, like: step, triangular, Mexican hat etc. 
The main requirements are: 1) $h_{k^*,k}(t)$ is symmetrical, and the maximum is obtained for $k=k^*$; 2) $h_{k^*,k}(t)$ decreases when $\Vert u_{k^*} -u_k \Vert$ increases. 

The learning rate $\alpha(t)$ and neighborhood size $\sigma(t)$ are usually decreased in time using an annealing scheme:
\begin{align}
\alpha(t+1) &= \alpha(t) \theta_{\alpha}, \\
\sigma(t+1) &= \sigma(t) \theta_{\sigma}, 
\end{align}
where $0<\theta_{\alpha}<1$ and $0<\theta_{\sigma}<1$ are the annealing scaling constants, while $\alpha(0)$ and $\sigma(0)$ are the initial values.
 
The above algorithm is repeated until some preset condition is met, for example until the change in the neurons weight is bellow a threshold $0<\varepsilon<1$:
\begin{equation}
\frac{1}{K} \sqrt{\sum_k \Vert u^{(k)}(t+1) - u^{(k)}(t) \Vert^2} \leq \varepsilon.
\end{equation}

Before training the weights of the SOM neurons must be initialized. The initialization is usually done using one of the 
following methods \cite{key-5}:
\begin{enumerate}
\item Independent random values.
\item Randomly drawn samples from the input data.
\item Principal components of the input data. 
\end{enumerate}

The above online algorithm is stochastic, and therefore the results depend on the order in which the input samples are presented to the SOM. 
A deterministic version of the SOM algorithm is the batch SOM, which instead of using a single random data point for each iteration, it uses the entire dataset \cite{key-6}. 
The batch update step is given by:
\begin{equation}
u^{(k)} = \frac{\sum_j n_j h_{j,k}\overline{x}^{(j)}}{\sum_j n_j h_{j,k}}, k=0,1,...,K-1.
\end{equation}
Here, $n_j$ is the number of input samples that have selected the node $j$ as their "winning node", and $\overline{x}^{(j)}$ is the average of all these data points. 
The value $h_{j,k}$ is the value of the neighborhood function, measuring the distance from node $j$ to node $k$. We should note that the batch algorithm 
does not require a learning rate, also its convergence is much faster than the online algorithm. 

An important aspect of SOMs is data visualization. A trained SOM consists of a 2-dimensional structure that is organized according to the 
the proximity of the samples in the input space, due to the neighborhood function role in topology preservation. Typically the nodes are 
represented by the cells of a regular grid (rectangular or hexagonal), the color (or intensity) of each cell is proportional to the average distance in the 
input space to the neighbors of that node. This can be used to visualize and identify data clusters in the high-dimensional input space, if for example lighter colors indicate 
higher distance (lower density), and darker colors indicate smaller distance (higher density) \cite{key-5}. 

\section{Diffusion Self-Organizing Map on the Unit Hypersphere}

In our implementation the input data and the SOM neurons are normalized to the unit hypersphere, also we replace the distance with the dot product as a similarity measure. 
That is, we assume that $\Vert x^{(n)} \Vert = 1$, $n=0,1,...,N-1$, and $\Vert u^{(k)} \Vert = 1$, $k=0,1,...,K-1$. We observe that in this case we have:
\begin{equation}
\Vert x^{(n)}(t) - u^{(k)}(t) \Vert^2 = 2[1-(x^{(n)}(t))^Tu^{(k)}(t)], 
\end{equation}
and therefore the competitive rule can be rewritten as:
\begin{equation}
k^* = \text{arg} \min_k \Vert x^{(n)}(t) - u^{(k)}(t) \Vert \Leftrightarrow k^* = \text{arg} \max_k (x^{(n)}(t))^T u_k(t)
\end{equation}

A second aspect is to replace the neighborhood function $h()$ with the solution of the diffusion equation, which is calculated as following. 
Let us assume that $k^*$ is the index of the winning neuron, and $(i^*,j^*)$ is its corresponding pair of indices on the SOM grid. 
We consider the matrix: 
\begin{equation}
h_{i,j}(0) =   \begin{cases}
   1  & (i,j) = (i^*,j^*) \\
   0 & (i,j) \neq (i^*,j^*)
  \end{cases}
\end{equation}
and we calculate the diffusion equation solution using the finite-difference approach \cite{key-7}:
\begin{equation}
h_{i,j}(\tau+1) = h_{i,j}(\tau) + D \Delta \tau [h_{i+1,j}(\tau)+h_{i-1,j}(\tau)+h_{i,j+1}(\tau)+h_{i,j-1}(\tau)-4h_{i,j}(\tau)]/(\Delta x)^2
\end{equation}
where $\tau$ is the (diffusion) time step, and $D$ is the diffusion coefficient. We impose also toroidal conditions on the grid, 
that is $i\pm1 \equiv (i\pm1)\text{mod}L$ where $L=\sqrt{K}$ ($K=L^2$). Also, we assume $\Delta \tau=1$ and $\Delta x=1$. During the computation we keep $h_{i^*,j^*}=1$ for all time steps $\tau$. 
We observe that by increasing the number time of steps we practically increase the radius of the neighborhood (Fig. 1). 
Also, if the number of steps $T$ is fixed, then we only have to compute the diffusion solution once for $(i^*,j^*)=(0,0)$, 
because we can simply shift the matrix such that the "winning neuron" is in the desired position $(i^*,j^*)=(i,j)$ (in numpy this can be done using the "$\text{roll}$" function). 

\begin{figure}[t!]
\centering
\subfigure[]{\includegraphics[width=6cm]{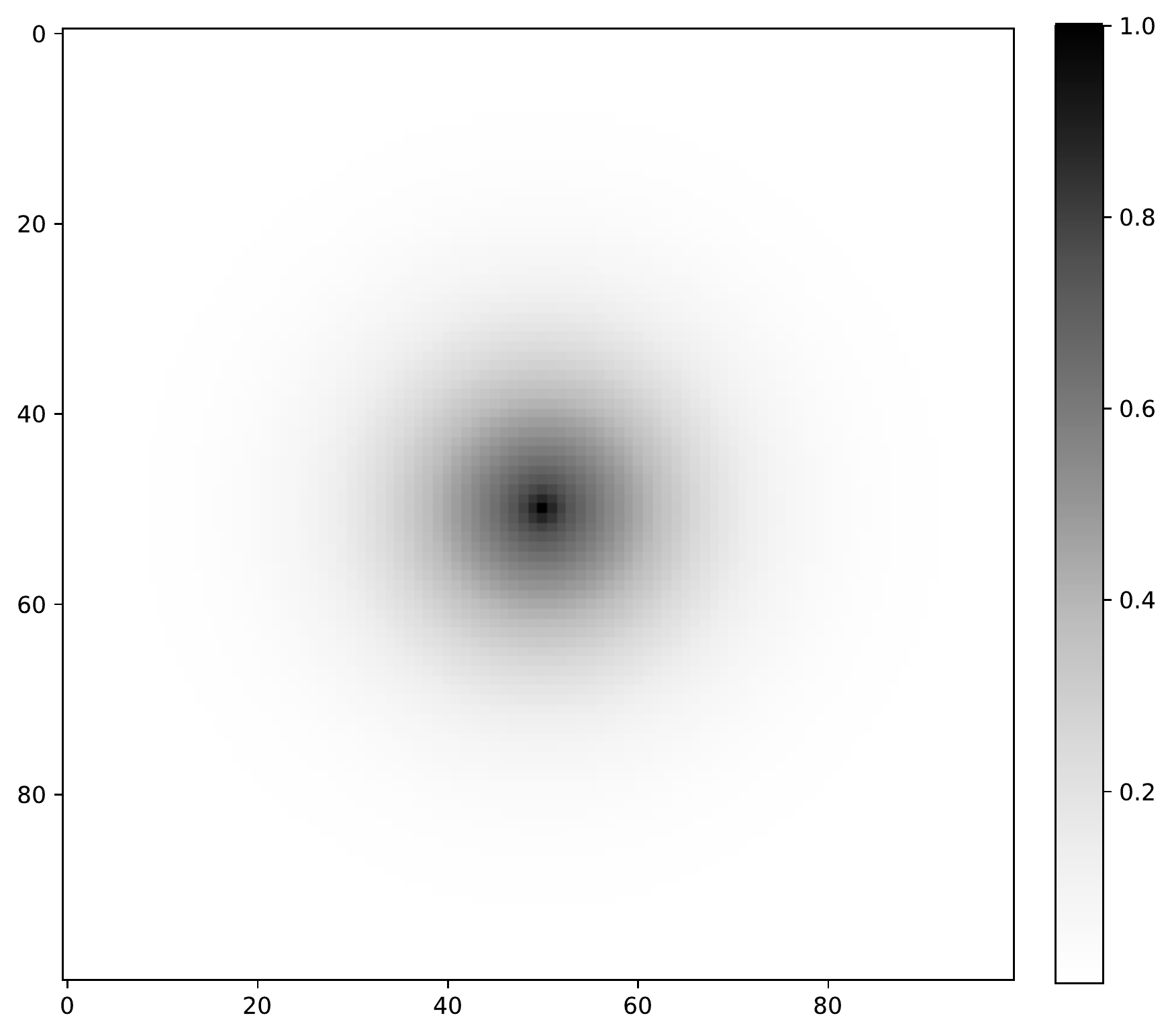}}
\qquad
\subfigure[]{\includegraphics[width=6cm]{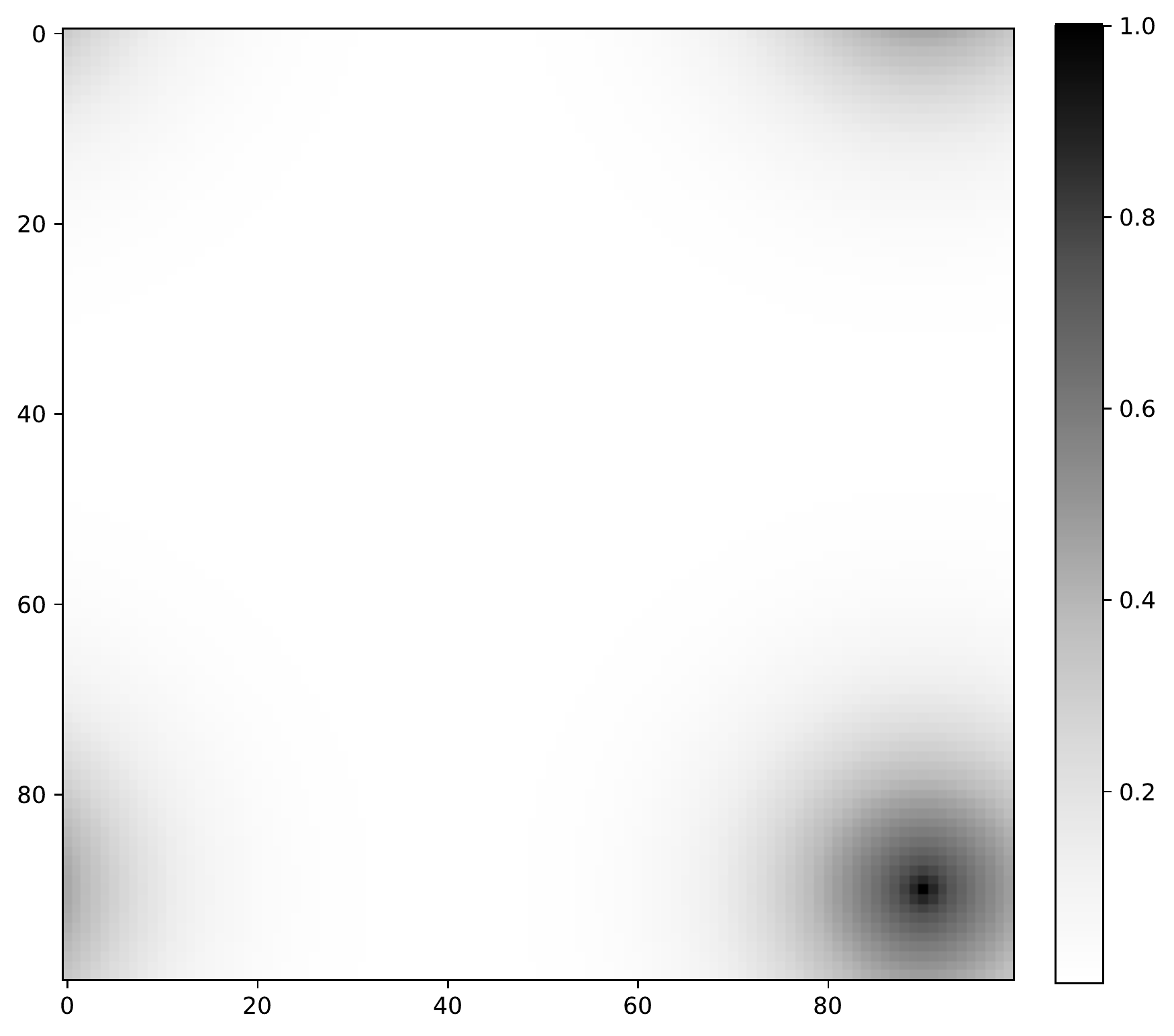}}
\caption{The solution of the diffusion equation for two different winning neurons ($L=100$).}
\end{figure}

For a given $h$ matrix we can formulate the following algorithm:

\begin{enumerate}
\item Initialize $0<\varepsilon<1$, $\delta \leftarrow 1$, $L=\sqrt{K}$.
\item Compute the diffusion matrix $h \leftarrow \text{diffusion}(L,D,T)$.
\item Randomly initialize the matrix $u  =[u^{(k)}]$ of $K$ neurons, such that each neuron is a normalized row $u^{(k)} \in \mathbb{R}^d$, $u^{(k)} \leftarrow u^{(k)}/\Vert u^{(k)} \Vert$, $k=0,1,...,K-1$.
\item While $\varepsilon \leq \delta$ compute:
\begin{equation}
v \leftarrow u
\end{equation}
\begin{equation}
r \leftarrow xu^T
\end{equation}
\begin{equation}
c \leftarrow \text{arg} \max_k r
\end{equation}
\begin{equation}
r^{(n)} = f(h,\lfloor c_n/L \rfloor, c[n]-L \lfloor c_n/L \rfloor), n=0,1,...,N-1.
\end{equation}
\begin{equation}
u \leftarrow r^Tx
\end{equation}
\begin{equation}
u^{(k)} = u^{(k)}/\Vert u^{(k)} \Vert, k=0,1,...,K-1.
\end{equation}
\begin{equation}
\delta = 1 - \frac{1}{K} \sum_k (u^{(k)})^T v^{(k)}
\end{equation}
\end{enumerate}

The algorithm parameters are $\varepsilon$, $\delta$ and $L$. Here, $\varepsilon$ is the smallest accepted error between two consecutive estimations of the SOM neurons, $\delta$ is the value of the time dependent error between two consecutive estimations of the SOM neurons. So, the algorithm runs until $\varepsilon \leq \delta$. $L$ is the length of the side of the square grid of SOM neurons, $K=L^2$. 
The next step is to compute the diffusion matrix $h$ with the parameters $L$, $D$ and $T$ (the number of diffusion time steps).
The neurons can be initialized using the same approach as for the standard SOM algorithm, in this particular case we just use random initialization. 

In (12) we make a copy of the neurons (for future error estimation). In (13) we compute the dot product between the input data $x$ and the SOM neurons $u$. 
We then compute the index of the "winning neurons" for each input sample (14). In the next step (15) we replace each row $r^{(n)}$ in $r$ with the shifted and flattened version 
of the diffusion matrix $h$, this operation is provided by the function $f(h,i,j)$ (here $\lfloor \cdot \rfloor$ is the floor function). 
With the new $r$ we can compute the new values of the SOM neurons $u$ in (16), which are then normalized in (17). Finally in (18), the "alignment" error $\delta$ between the previous SOM neurons and the current 
SOM neurons is computed. If $\delta < \varepsilon$ the algorithm stops, otherwise it repeats.

\section{MNIST dataset}

In order to illustrate our approach we consider the well known MNIST data set, which is a large database of handwritten digits (0,1,...,9), containing 60,000 training images 
and 10,000 testing images \cite{key-8,key-9}. These are monochrome images with an intensity in the interval $[0,255]$ and the size of $28 \times 28$ pixels. 
The MNIST data set is probably the most frequently used benchmark in image recognition. 
We normalize the images as following:
\begin{align}
x_n &\leftarrow x_n - \langle x_n \rangle \\
x_n &\leftarrow x_n/\Vert x_n \Vert.
\end{align}

\begin{figure}[t!]
\centering
\subfigure[]{\includegraphics[width=6cm]{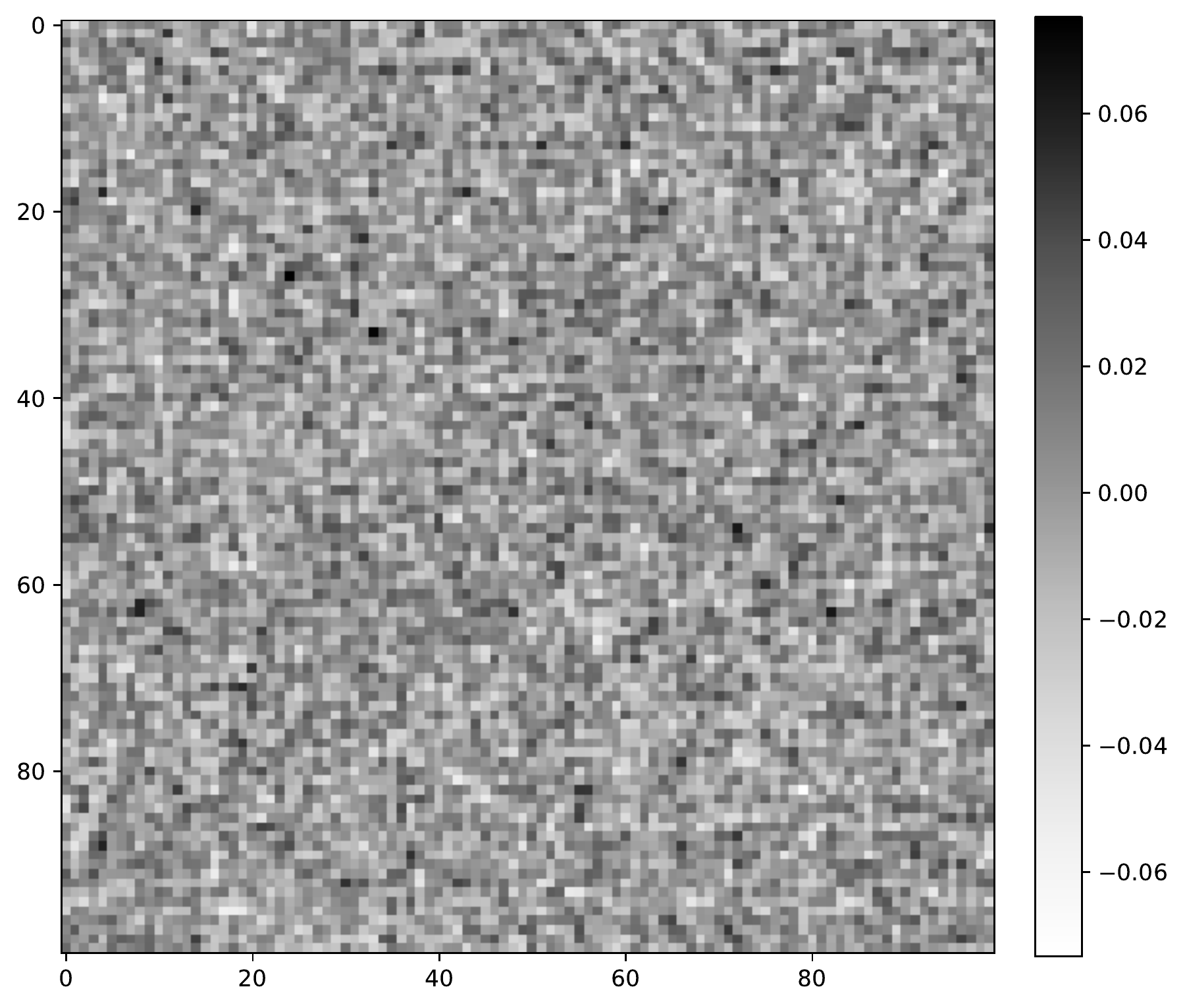}}
\qquad
\subfigure[]{\includegraphics[width=6cm]{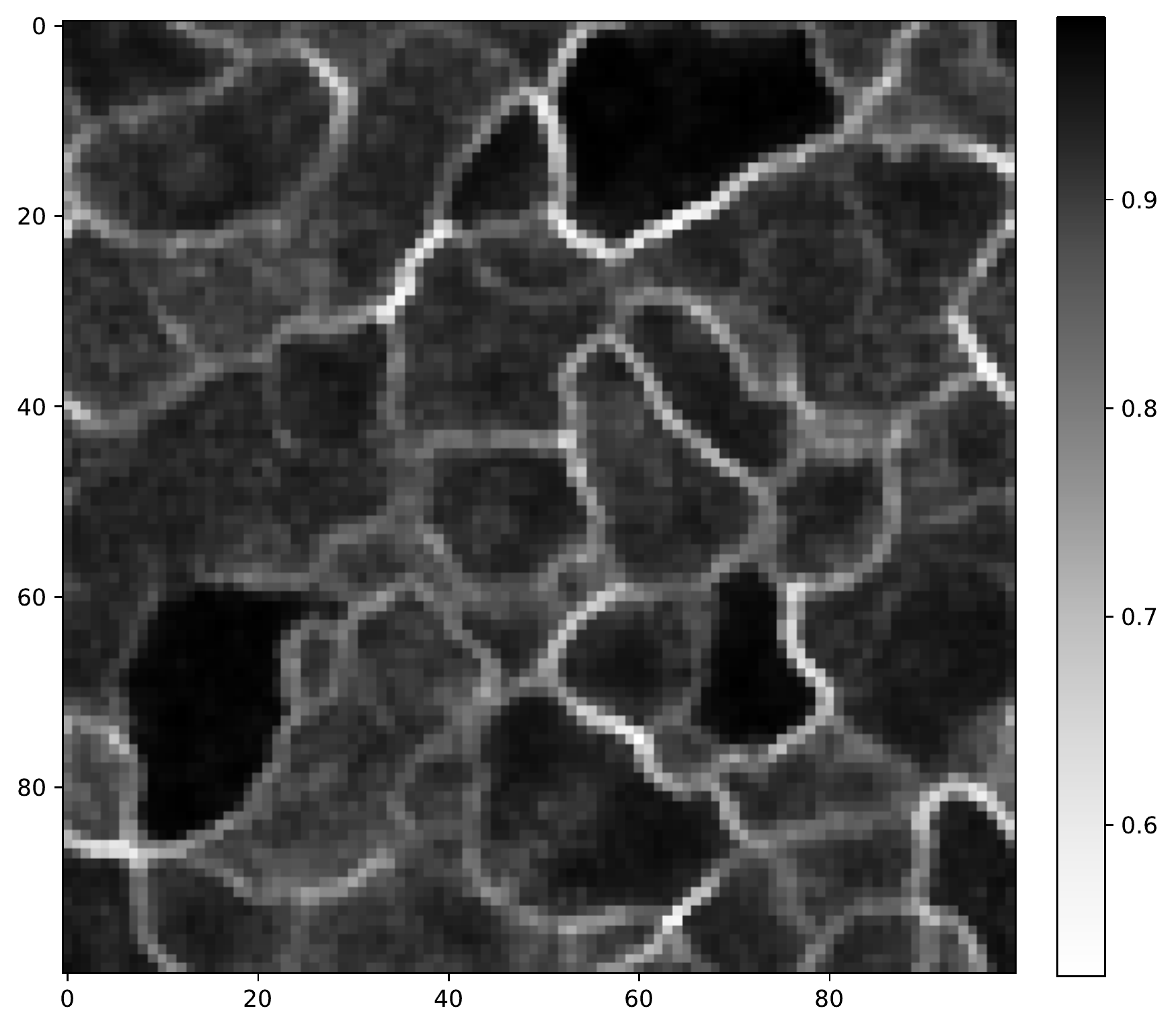}}
\caption{D-SOM for MNIST data ($L=100$): (a) untrained D-SOM (b) trained D-SOM.}
\end{figure}

In Fig. 2 we give the resulted SOM visualization for the following parameters: $L=10^2$, $K=10^4$, $D=0.25$. 
Fig. 2(a) shows the initial correlation of the neurons, while Fig. 2(b) shows the final correlation, after training the SOM. 
One can see clearly how the domain valleys  
with lower correlation (light grey) are formed between darker regions occupied by neurons associated with different classes. 

\begin{figure}[h!]
\centering
\subfigure[]{\includegraphics[width=5.5cm]{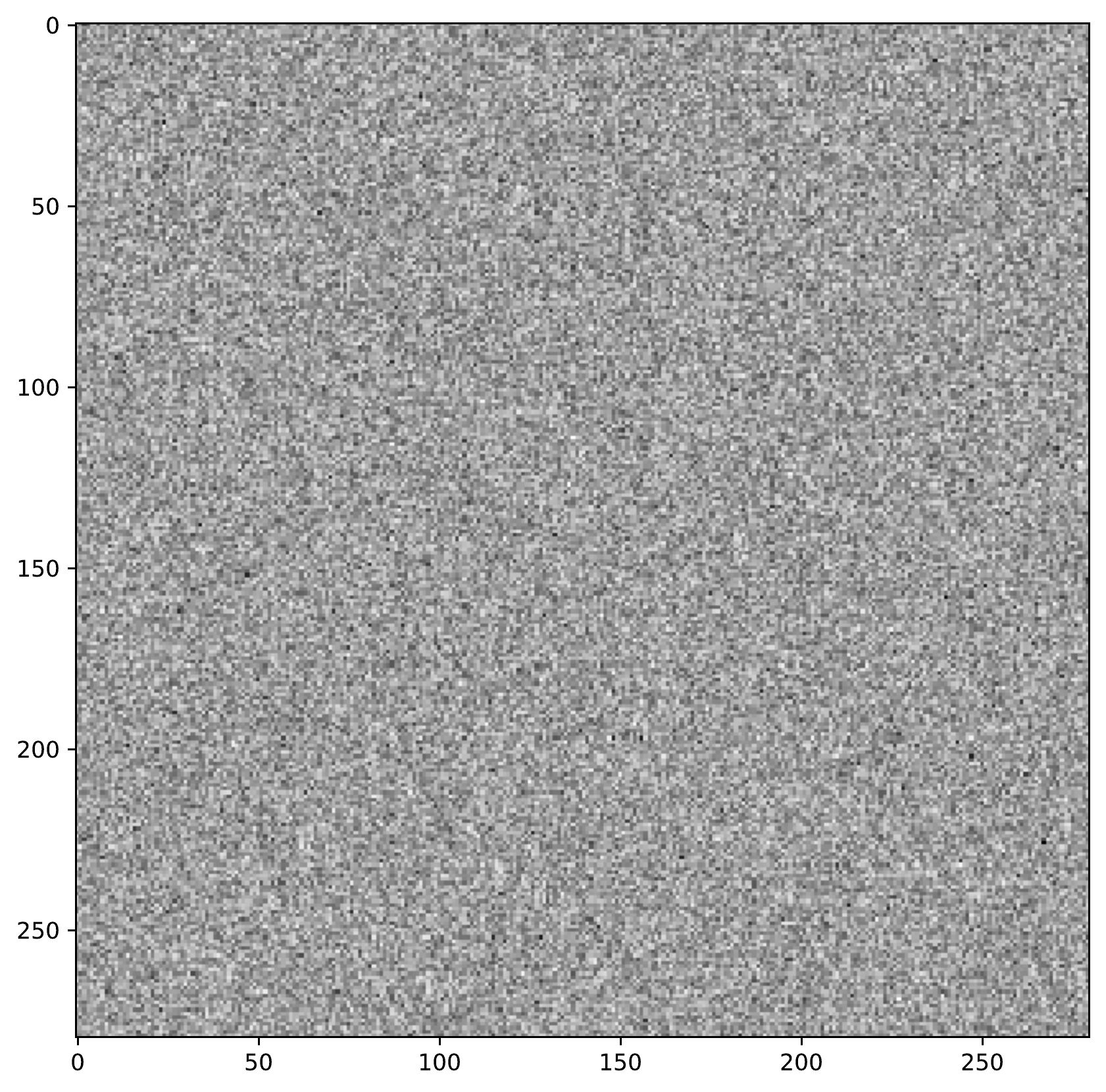}}
\qquad
\subfigure[]{\includegraphics[width=5.5cm]{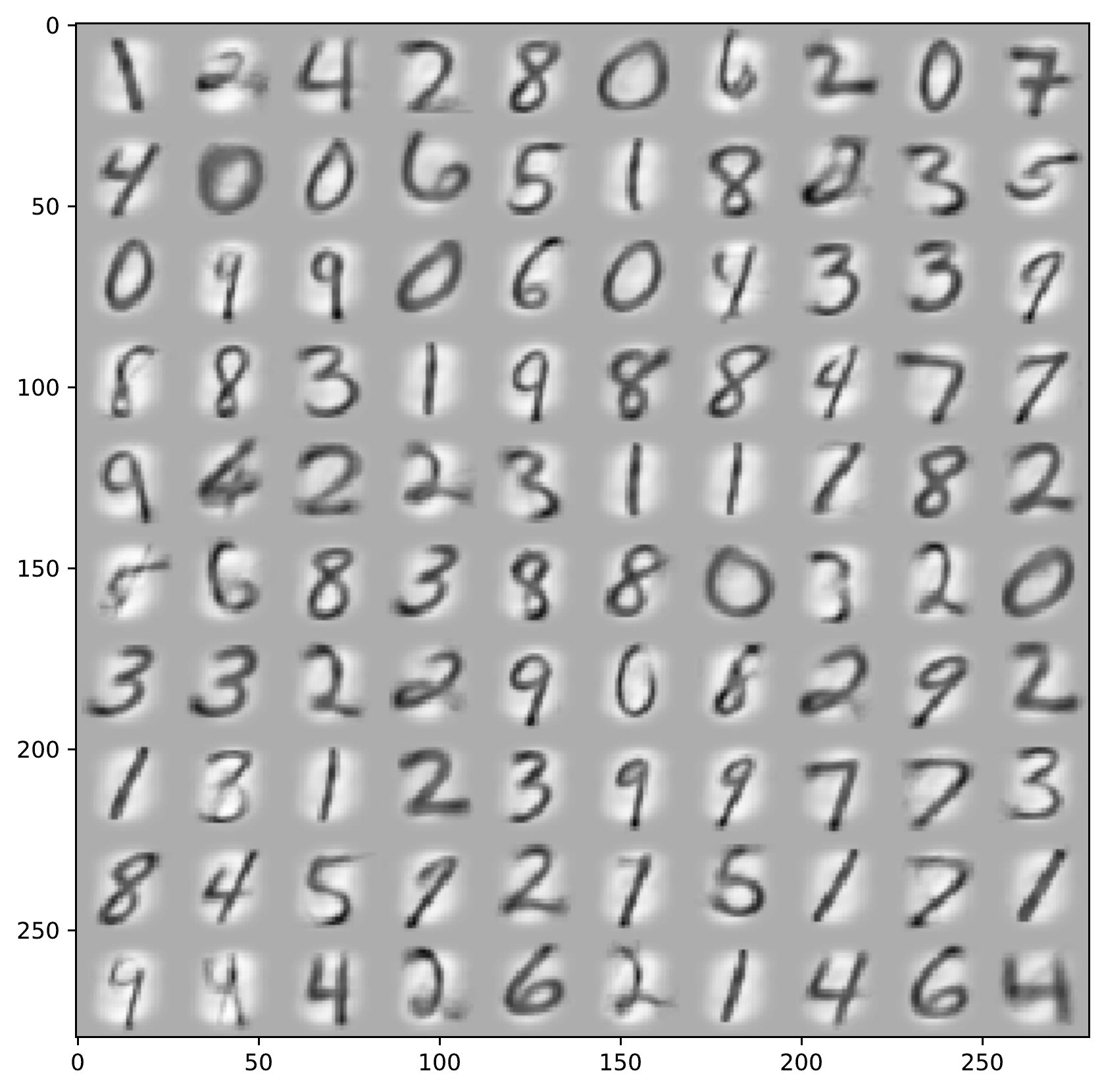}}
\caption{D-SOM neurons: (a) before training (b) after training.}
\end{figure}

In Fig. 3 we also show a random sample of 100 SOM neurons, before and after training. One can see that each neuron is actually learning 
a class of digits. 
While the SOM is primarily used for visualization, it can also be used for classification. 
In this case with each neuron $u^{(k)}$ we associate the class of the input sample that has the highest correlation (dot product). 
In Fig. 4 we show the SOM regions corresponding to each class [0,1,...,9]. 
In order to classify a new sample, one can just compute the dot product between this sample and all the SOM vectors, and the corresponding 
class will be the class of the neuron with the highest correlation. This, simple procedure provides an accuracy better than $95\%$, which is not bad 
considering its simplicity.

\begin{figure}[t!]
\centering
\subfigure[0]{\includegraphics[width=3cm]{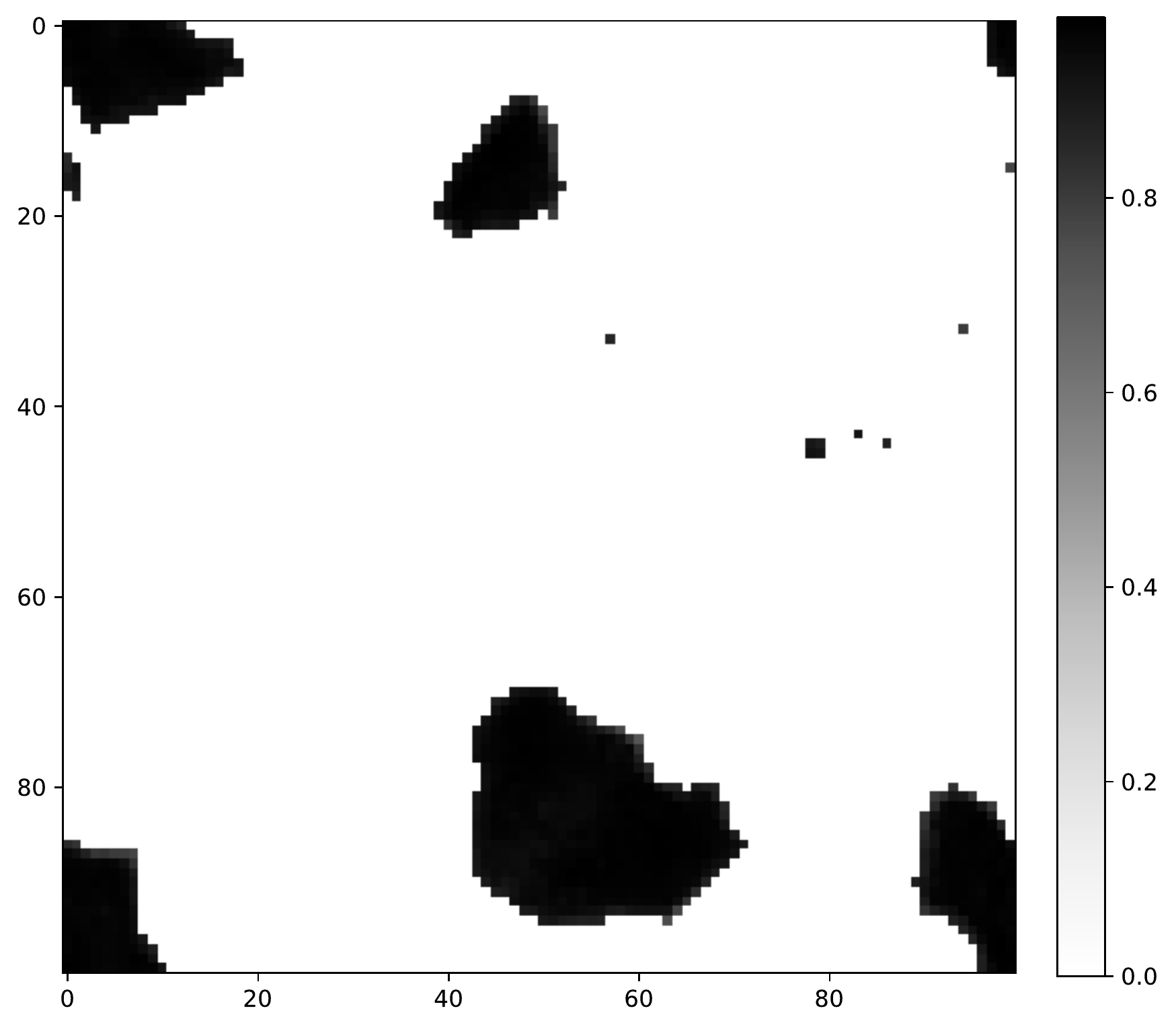}}
\qquad
\subfigure[1]{\includegraphics[width=3cm]{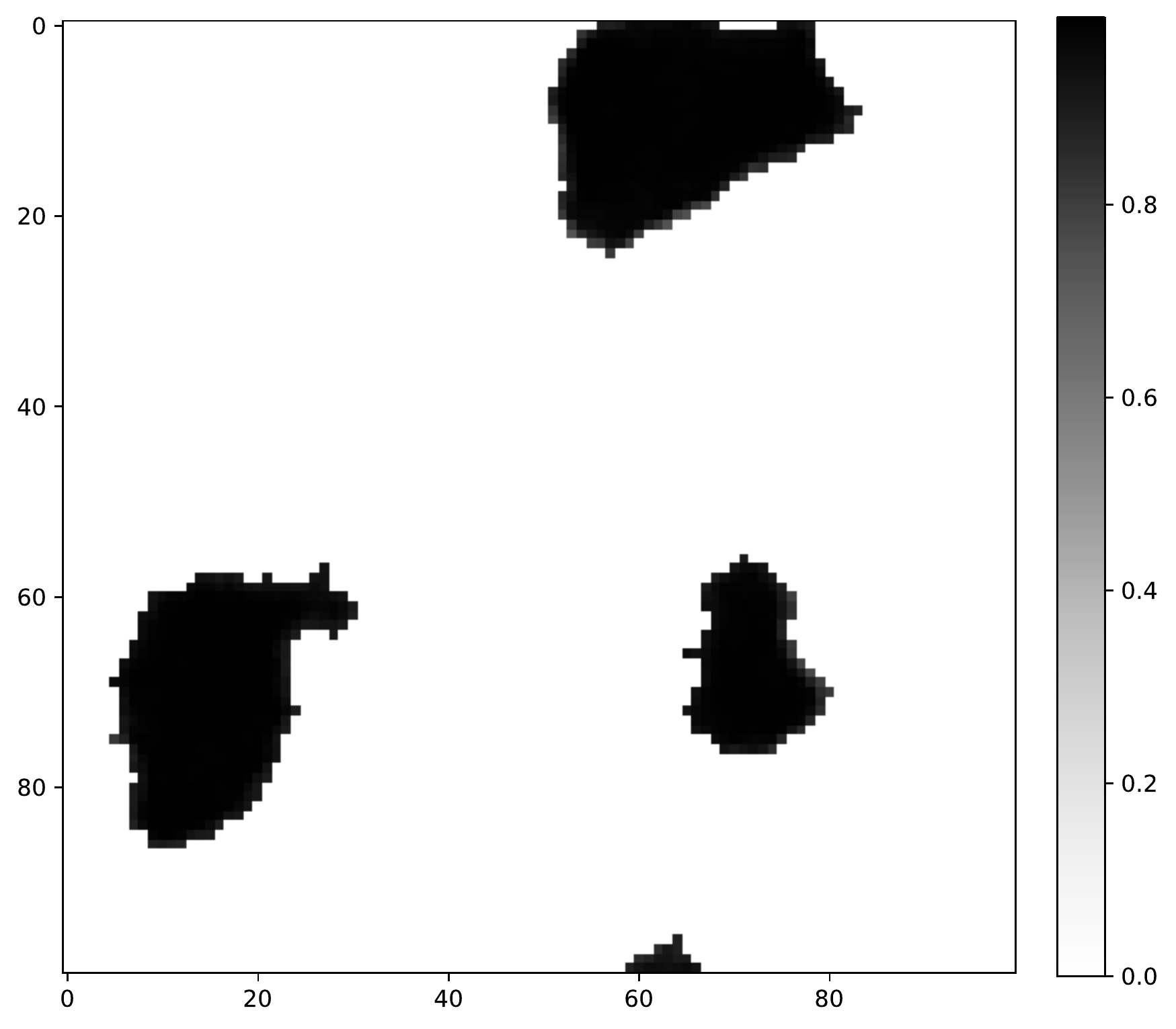}}
\qquad
\subfigure[2]{\includegraphics[width=3cm]{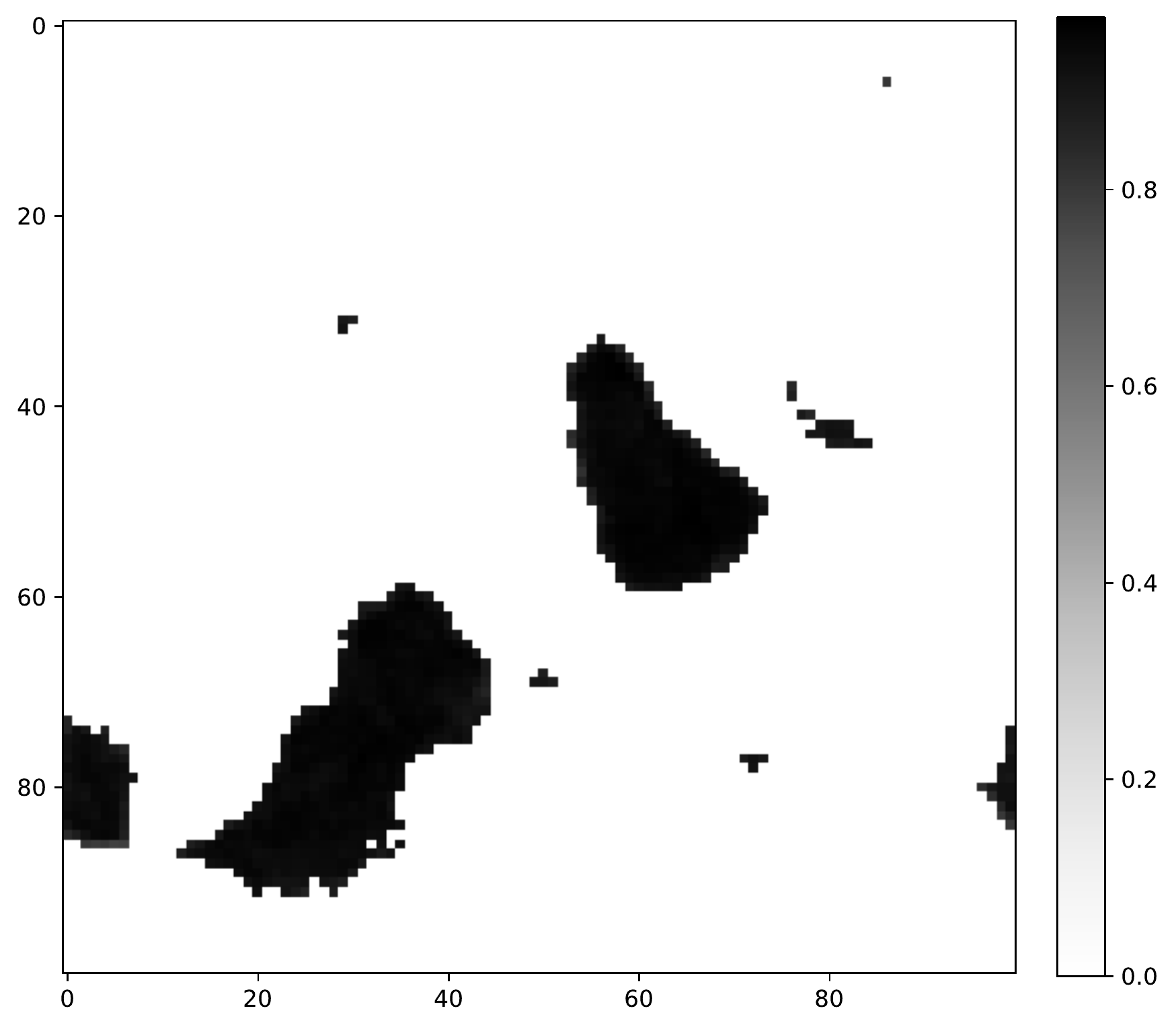}}
\qquad
\subfigure[3]{\includegraphics[width=3cm]{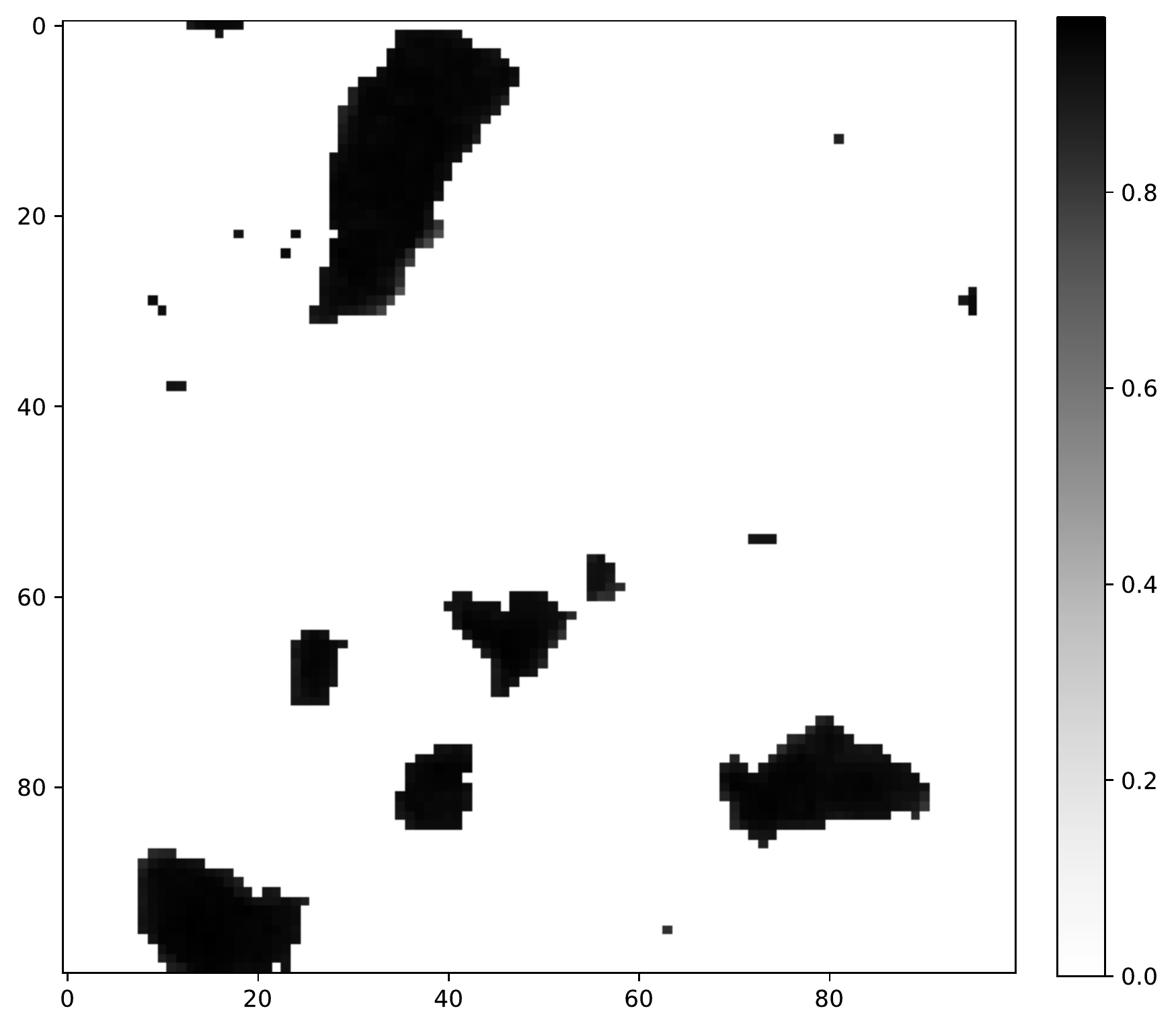}}

\subfigure[4]{\includegraphics[width=3cm]{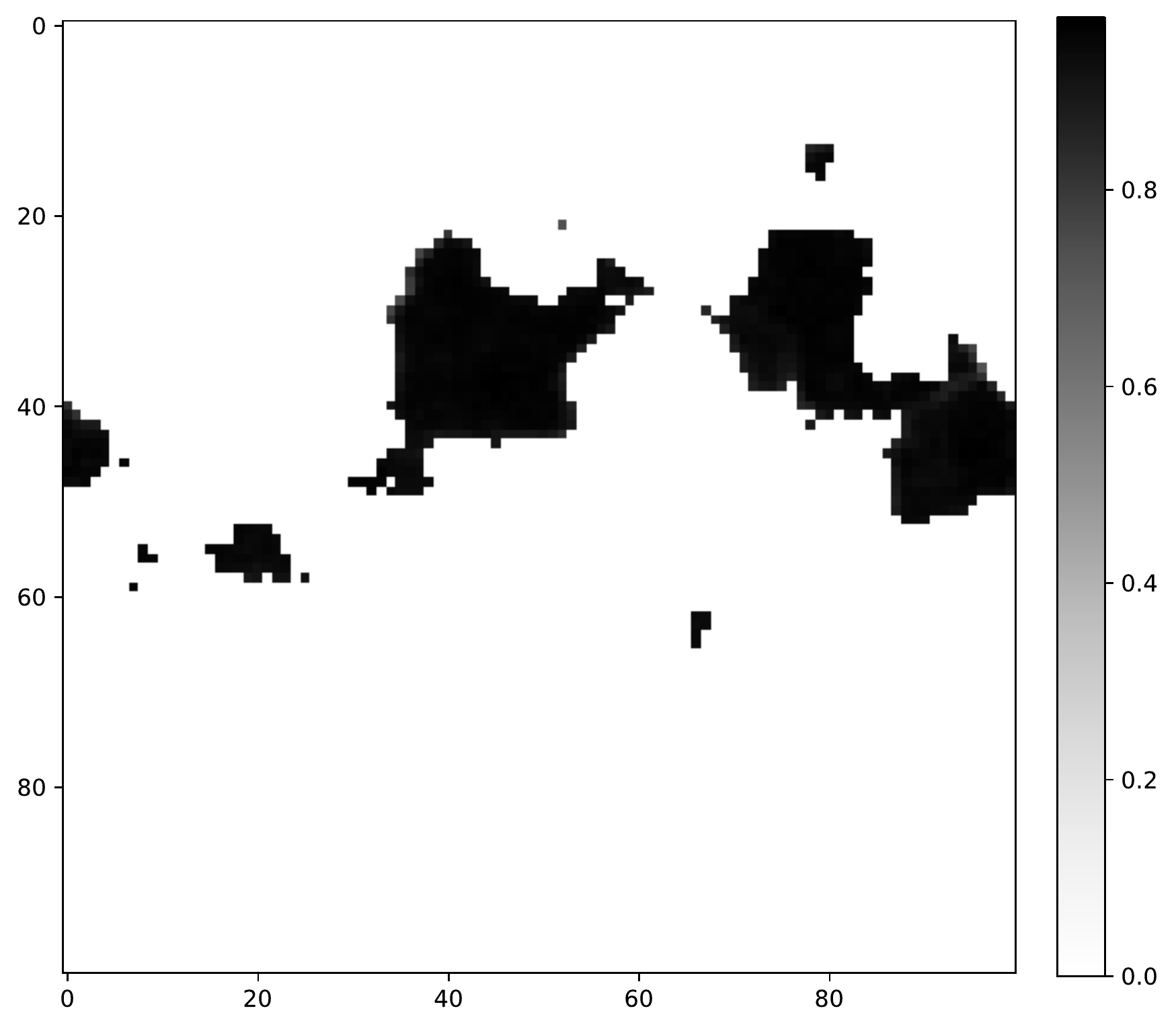}}
\qquad
\subfigure[5]{\includegraphics[width=3cm]{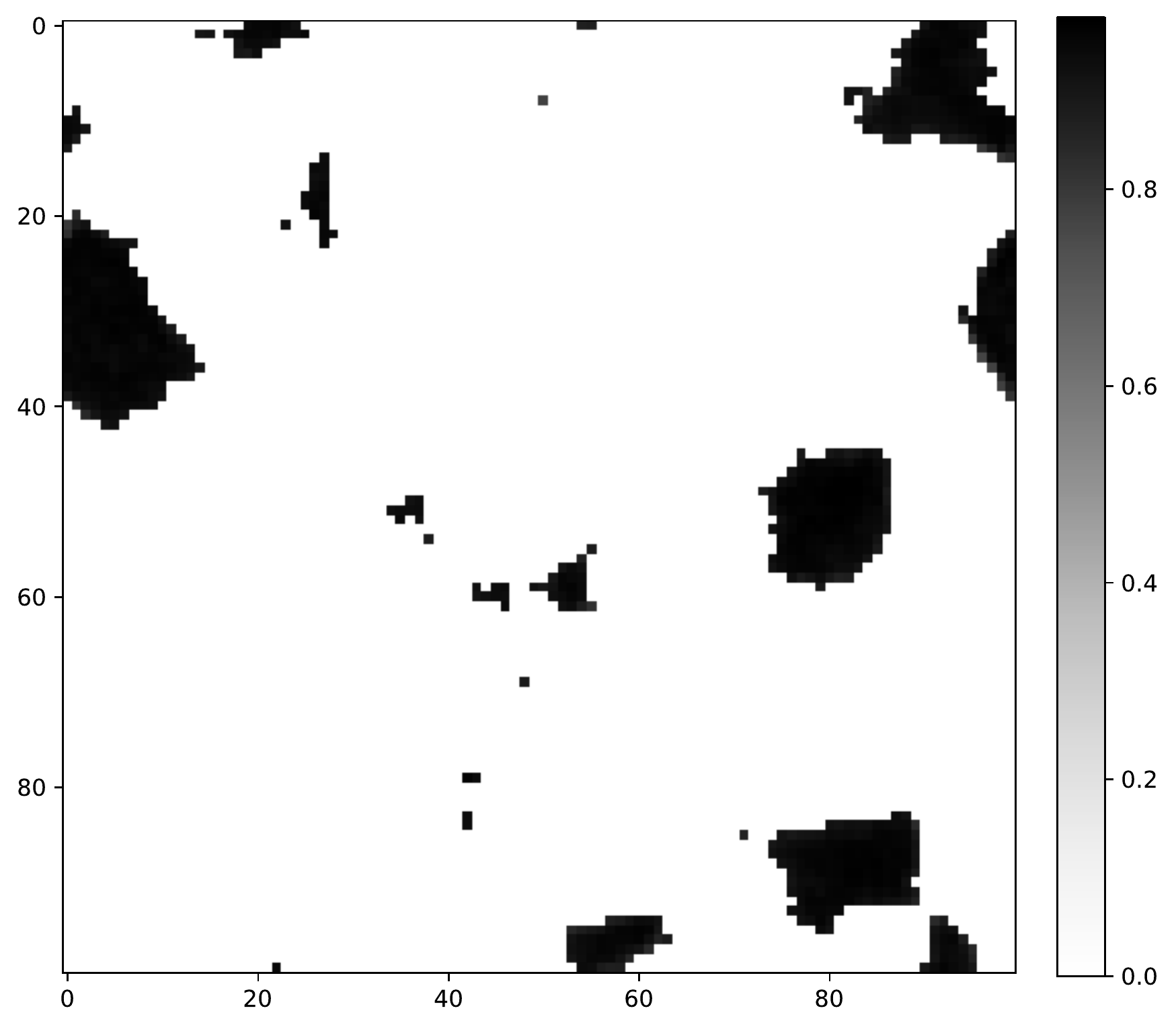}}
\qquad
\subfigure[6]{\includegraphics[width=3cm]{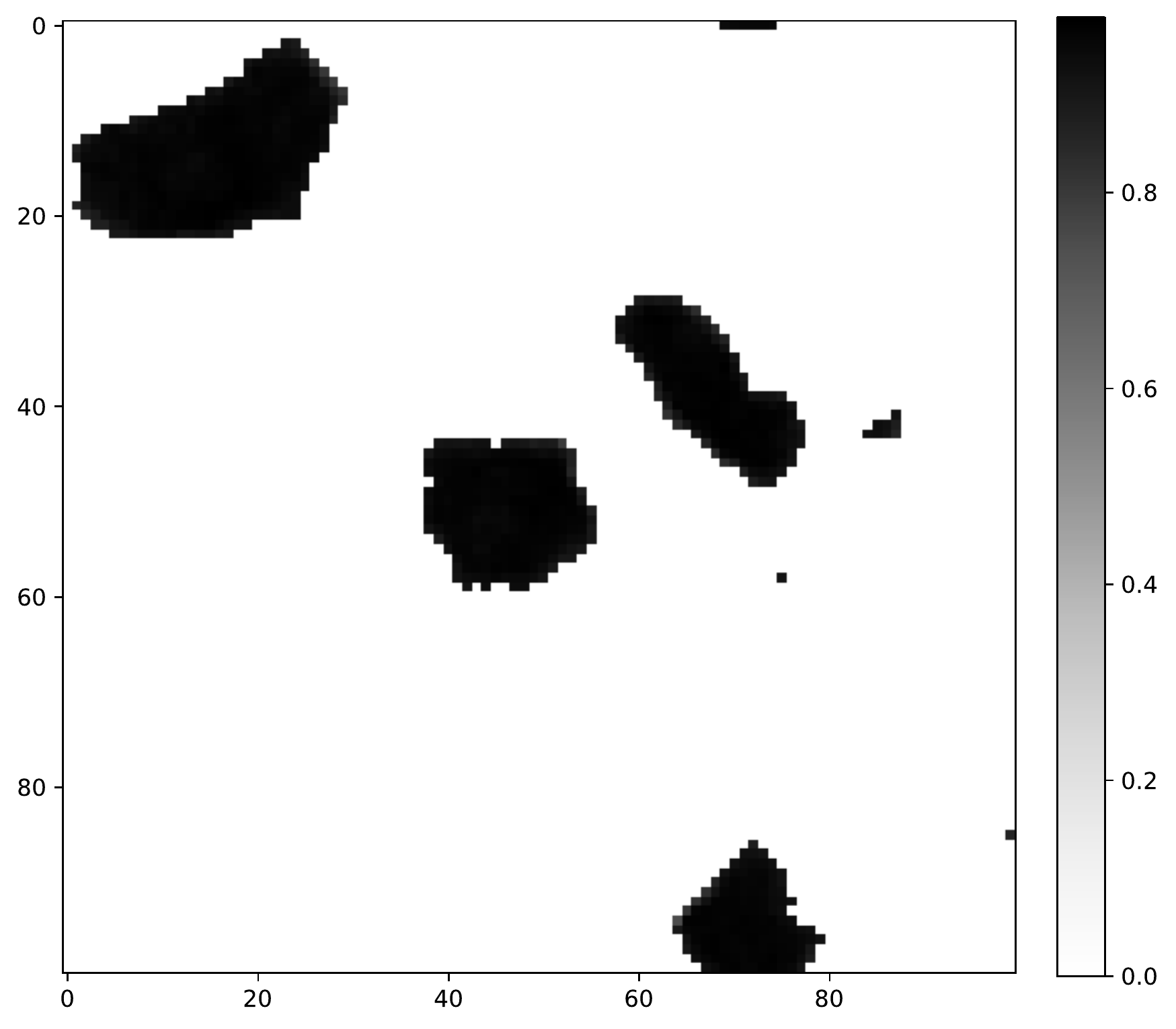}}
\qquad
\subfigure[7]{\includegraphics[width=3cm]{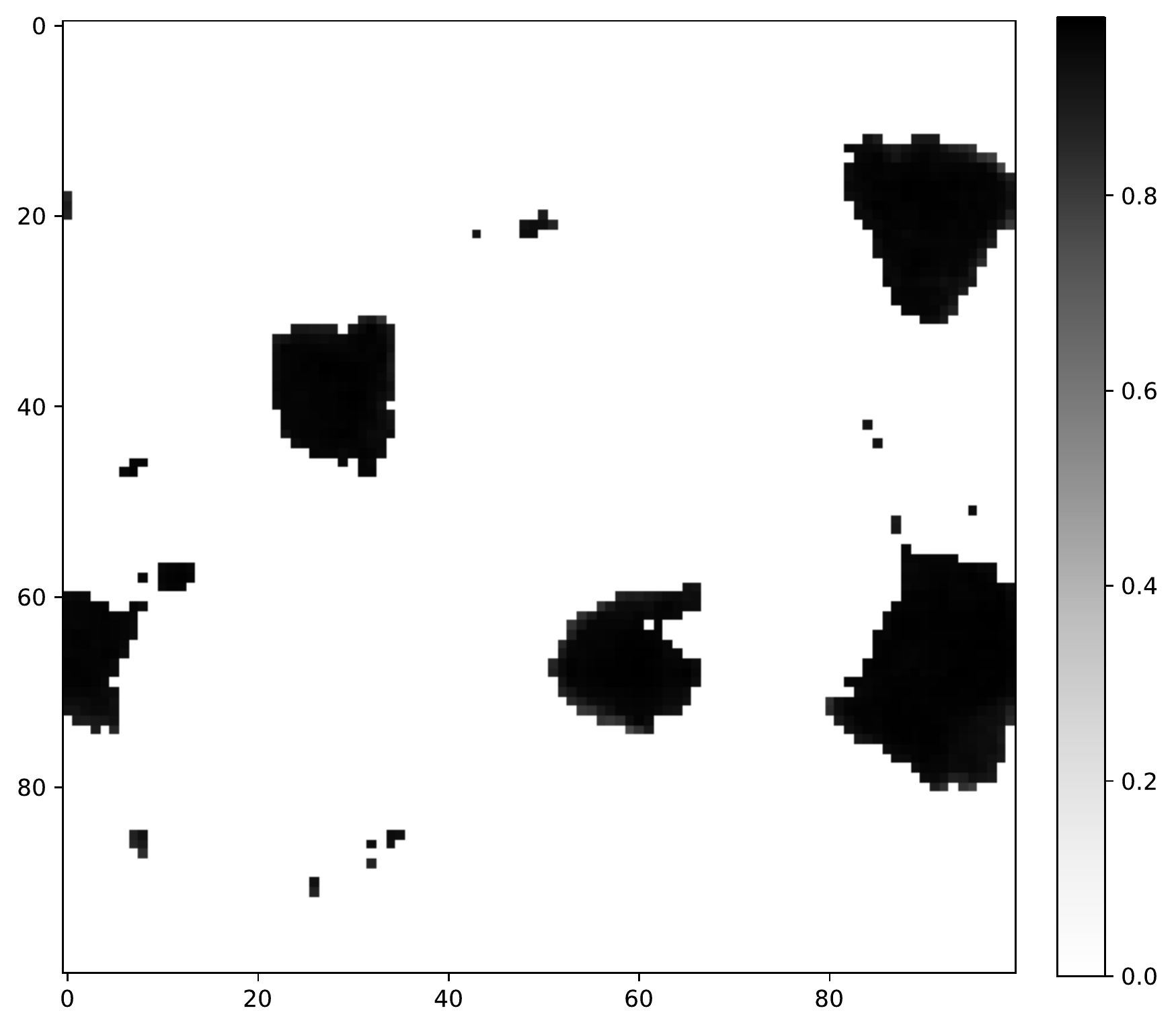}}

\subfigure[8]{\includegraphics[width=3cm]{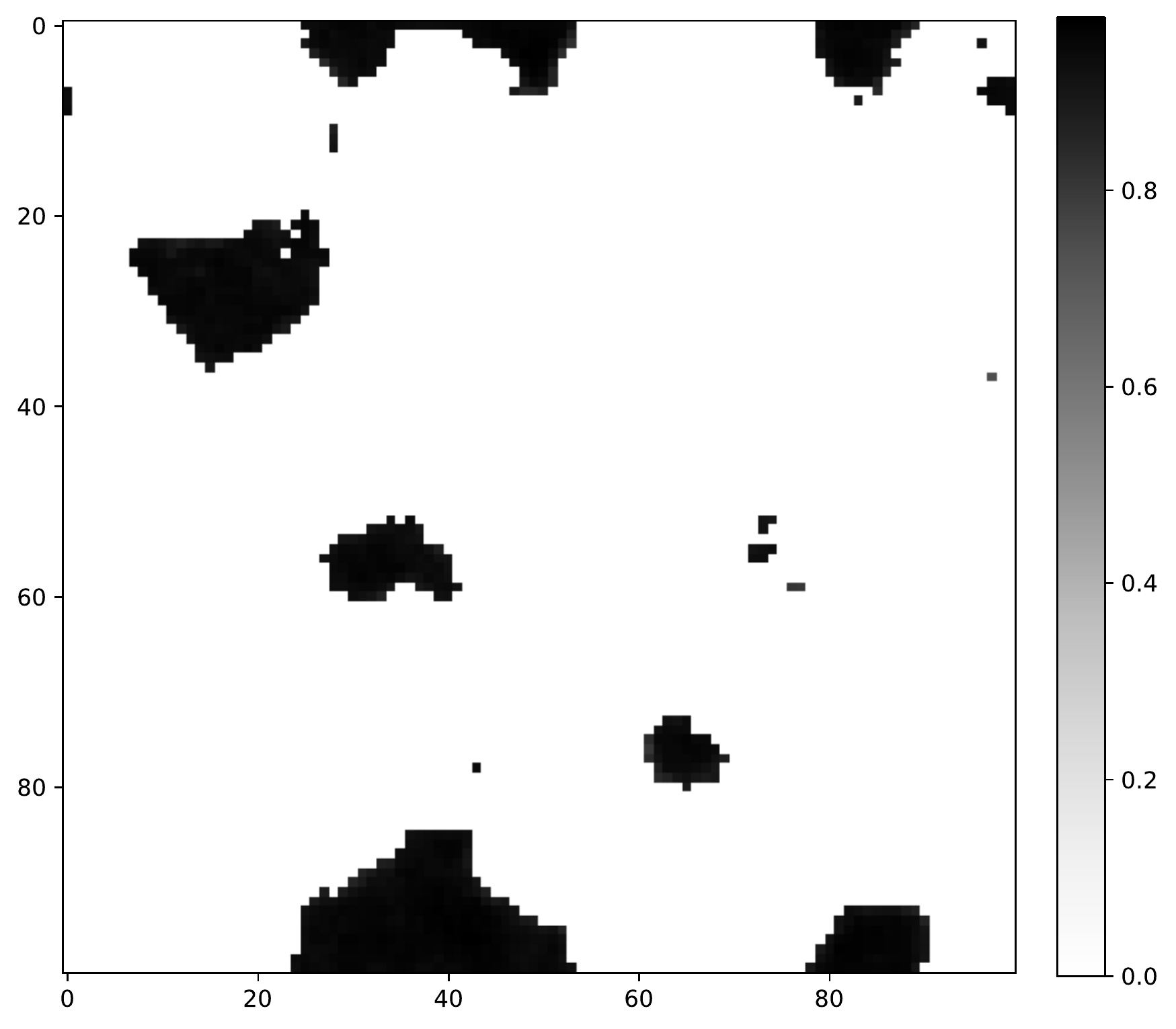}}
\qquad
\subfigure[9]{\includegraphics[width=3cm]{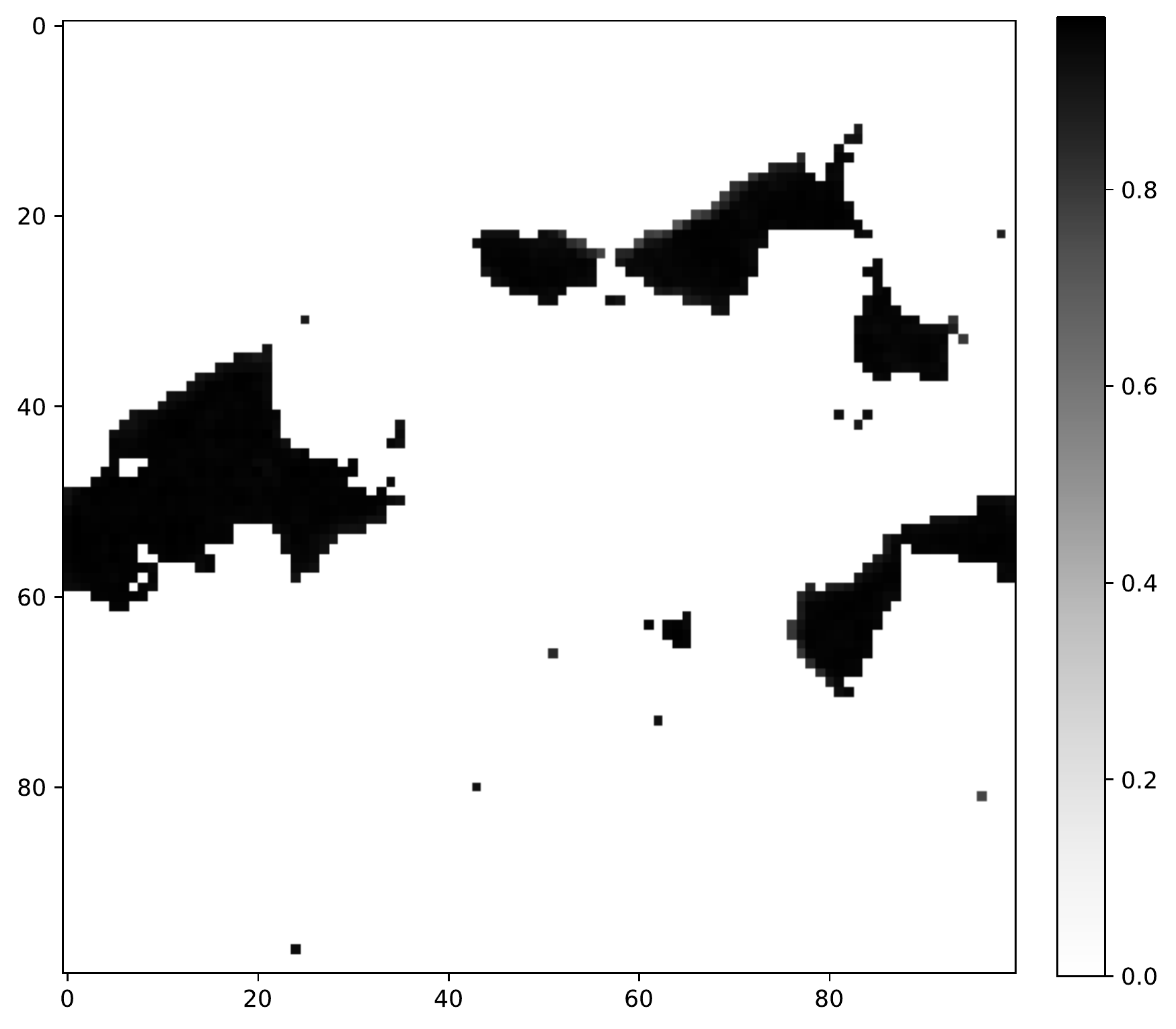}}

\caption{D-SOM class regions for MNIST data.}
\end{figure}

\section{Numerical implementation}

The D-SOM can be implemented efficiently in python numpy, by exploiting the parallelism provided by the Intel's MKL BLAS library. 
Below we give the implementation for the MNIST data set:

\footnotesize
\begin{lstlisting}[language=Python]
import numpy as np
import matplotlib.pyplot as plt

def read_data(imagefile,labelfile,N,M):
    x = np.zeros((N,M),dtype=np.float32)
    images = open(imagefile,'rb')
    images.read(16)  # skip the magic_number
    for n in range(N):
        x[n,:] = np.frombuffer(images.read(M),dtype='uint8').astype(np.float32)
    images.close()
    x = x - np.mean(x,axis=0)
    a = np.linalg.norm(x,axis=1)
    x = x/a[:,None]
    labels = open(labelfile,'rb')
    labels.read(8)  # skip the magic_number
    xl = np.frombuffer(labels.read(N), dtype='uint8')
    labels.close()
    return (x, xl)

def diffusion(L,T):
    h = np.zeros((L,L),dtype=np.float32)
    h[0,0],D = np.float32(1.0),np.float32(0.25)
    for t in range(T):
        g = np.vstack((h[1:,:],h[0,:]))
        g += np.vstack((h[L-1,:],h[:L-1,:]))
        g += np.hstack((h[:,1:],h[:,0].reshape(L,1)))
        g += np.hstack((h[:,L-1].reshape(L,1),h[:,:L-1]))
        h = D*g
        h[0,0] = np.float32(1.0)
    return h

def dsom(x,u):
    v,(N,M),K = u,x.shape,len(u)
    L = int(np.sqrt(K))
    for t in range(L//2,1,-1):
        h = diffusion(L,t)
        delta,eta = 1,0
        while delta > 1e-6 and delta != eta:
            eta = delta
            r = np.dot(x,u.T)
            c = np.argmax(r,axis=1)
            for n in range(N):
                r[n,:] = np.roll(h,(c[n]//L,c[n]-L*(c[n]//L)),
                                 axis=(0,1)).flatten()
            u = np.dot(r.T,x)
            a = np.linalg.norm(u,axis=1)
            u = u/a[:,None]
            delta,v = 1 - np.sum(u*v)/K,u
            print("t=",t,"delta=",delta)
    return u

def som_correlation(u):
    (K,M) = u.shape
    L = int(np.sqrt(K))
    w = np.zeros((L,L),dtype=np.float32)
    for i in range(L):
        for j in range(L):
            n = i*L + j
            m = (n - L)%K
            w[i,j] += np.dot(u[n,:],u[m,:])
            m = (n - 1)%K
            w[i,j] += np.dot(u[n,:],u[m,:])
            m = (n + 1)%K
            w[i,j] += np.dot(u[n,:],u[m,:])
            m = (n + L)%K
            w[i,j] += np.dot(u[n,:],u[m,:])
            w[i,j] /= 4
    return w

def init_som(K,M):
    u = np.random.randn(K,M).astype(np.float32)
    for k in range(K):
        u[k,:] = u[k,:]/np.linalg.norm(u[k,:])
    return u

if __name__ == '__main__':
    np.random.seed(12345)
    print("Read data")
    (x, xl) = read_data("./data/train-images.idx3-ubyte", 
                        "./data/train-labels.idx1-ubyte", 60000, 784)
    (N,M) = x.shape
    K = 10000 # number of SOM neurons
    u = init_som(K,M)
    print("Train SOM")
    u = dsom(x,u)
    np.savetxt("som-u.csv",u,delimiter=",",fmt='%f')
    print("Visualize SOM")
    w = som_correlation(u)
    fig = plt.figure(figsize=(8,8))
    hmap = plt.imshow(w, cmap='Greys', interpolation='nearest')
    plt.colorbar(hmap,fraction=0.046, pad=0.04)
    fig.savefig("som.pdf",bbox_inches="tight")
\end{lstlisting}
\normalsize

The implementation is also using 32 bit float numbers to minimize the memory footprint. 
The "read\_data" function reads the data "x" and normalizes it. The number of neurons in SOM is set to $K=10^4$ and they are initialized randomly in "init\_som". 
The "dsom" function implements the Diffusion-SOM algorithm. As input it takes the data array "x" and the initialized SOM vectors "u". 
The diffusion time is varied from $L/2$ to 1, such that initially the neighborhood is large (facilitating long range interactions between neurons) with a radius of $L/2$, and then it shrinks to the the first neighbors only. 
For each diffusion time one computes a diffusion matrix "h". Here the diffusion coefficient is set to D=0.25. Then for each "h" one applies the algorithm described by (12)-(18). The algorithm stops when 
"delta" becomes smaller or equal to $\varepsilon = 10^{-6}$ or if it stops converging. 

The SOM vectors are visualized by a square image where each neuron is associated to a pixel, with the intensity corresponding to the average correlation (dot product) between that 
neuron and its immediate four neighbors, which is computed by the function "som\_correlation". Finally, the resulted map is saved in "som.pdf".

\section*{Conclusion}
In this paper we have presented a version of SOM that is implemented on the unit hypersphere, and uses the dot product as a similarity measure. 
The nodes of the network also use diffusion to communicate among them. This approach can be efficiently implemented using just linear algebra methods, and  
a python numpy version is provided. The method was illustrated using the MNIST dataset. We should also note that 
if the number diffusion time steps is reduced to one (no diffusion), then the D-SOM algorithm becomes equivalent to the K-Means algorithm described in \cite{key-10}.

\end{document}